\documentclass[conference,compsoc]{IEEEtran}
\ifCLASSOPTIONcompsoc
  \usepackage[nocompress]{cite}
\else
  \usepackage{cite}
\fi

\usepackage{graphicx}
\usepackage{amsmath}
\usepackage{amssymb}

\usepackage{booktabs}
\usepackage{array}

\usepackage{subcaption}

\usepackage{algorithm}
\usepackage{algpseudocode}

\usepackage{float}
\usepackage{placeins}

\usepackage{xcolor}
\usepackage{url}
\ifCLASSINFOpdf
\else
\fi
\begin{document}
%
\title{Digital Twins for Opinion Dynamics: \\ A Generative LLM Framework for Social Networks}

\author{\IEEEauthorblockN{Omran Berjawi}
\IEEEauthorblockA{Information Processing and Communication Laboratory (LTCI)\\
Institut Polytechnique de Paris, T\'el\'ecom Paris\\
Palaiseau, France\\
Email: omran.berjawi@telecom-paris.fr}
\and

\IEEEauthorblockN{Giuseppe Fenza}
\IEEEauthorblockA{ Management \& Information Systems (DISA-MIS)\\
University of Salerno\\
Fisciano, Italy\\
Email: gfenza@unisa.it}
\and
\IEEEauthorblockN{Rida Khatoun}
\IEEEauthorblockA{Information Processing and Communication Laboratory (LTCI)\\
Institut Polytechnique de Paris, T\'el\'ecom Paris\\
Palaiseau, France\\
Email: rida.khatoun@telecom-paris.fr}
\and

\IEEEauthorblockN{Sherali Zeadally}
\IEEEauthorblockA{College of Communication and Information, \\
University of Kentucky\\
Department of Electronic Engineering, \\
Kyung Hee University, South Korea\\
Email:szeadally@uky.edu}}


%


\maketitle

\begin{abstract}
The study of opinion dynamics in social networks is one of the key challenges in computational social science with direct relevance to understanding political polarization, misinformation, and health responses. Current approaches focus on simplified mathematical models that ignore linguistic and contextual factors related to belief updates or use Large Language Model (LLM)-based simulations that have not been validated against real data. We present a framework based on the concept of a digital twin to simulate opinion dynamics in social networks. The approach fills the gap by cloning a real-world Twitter network, assigns a set of attributes for agents (such as persona, emotions, centrality, stubbornness, and influence), and employs Mistral-7B to perform opinion update based on memory and social exposure. To evaluate the proposed approach, we validate it against two real Twitter datasets (COVID-19 discourse and U.S elections 2020). The results show that the capability of the proposed framework reproduces opinion trajectories and reduces individual prediction error by more than 50\% compared to the best-performing classical baseline (Mistral-7B achieves Mean Absolute Error (MAE) = 0.150 and 0.121 on the COVID-19 and US Election 2020 datasets, respectively). We observe similar improvements in structural alignment ($\Delta$r = 0.120 and 0.180) and polarization dynamics ($\Delta$Var = 0.106 and 0.115) on the two datasets, respectively. Additionally, the ablation studies confirm that agent attributes, memory, and social exposure all contribute to the framework's predictive fidelity in reproducing opinion trajectories, with agent attributes being the most critical contributor. Overall, our results demonstrate that grounding Mistral-7B within empirically cloned interaction networks produces a realistic simulation framework capable of reproducing complex social dynamics.

\end{abstract}

\IEEEpeerreviewmaketitle

\section{Introduction}
Social media has transformed public discussion into a system that impacts world events and outcomes. According to a study of over 0.5 billion tweets associated with vaccines, there is a strong link between Twitter discussions and the number of people who received COVID-19 vaccinations at each immunization stage \cite{nelson2024vaccine,2024multi}. Similarly, online anti-vaccine content has been associated with reduced vaccination likelihood, with some analyses suggesting unreported misinformation may be substantially more damaging than reported misinformation. Beyond public health, narratives related to the U.S elections in 2020 via social media played a pivotal role in creating political polarization unlike any other time in history ~\cite{cinelli2021echo, van2024political, metzler2024social}. Therefore, predicting how opinions change over time is crucial for understanding the process of collective behavior.

\subsection{Research gaps}
The dominant computational approaches for opinion dynamics modeling in social networks consider opinion change as the result of dynamical processes that occur within a network, where the opinion change process is mathematically formulated via deterministic or stochastic aggregation functions operating on static interaction graphs. Classic models include the DeGroot~\cite{degroot1974reaching} and Friedkin--Johnsen models~\cite{friedkin1990social}, together with bounded confidence models like Hegselmann--Krause~\cite{rainer2002opinion} and Deffuant--Weisbuch~\cite{deffuant2000mixing}, produce analytical frameworks which describe the mechanisms of influence spread, consensus achievement, and polarization. Despite their mathematical tractability and interpretability, these models suffer from a significant drawback: they essentially ignore the cognitive processes underlying belief change. In particular, they model opinion updates primarily through numerical averages of agents' opinion values, without considering the language context; for instance, opinions in social networks are expressed through natural language, where context, framing, and emotional tone influence their interpretation and subsequent belief updates.

The solution provided by Large Language Models (LLMs) offers a completely novel way to address this limitation(i.e., the inability of traditional models to capture the linguistic context of opinion change). This is accomplished with generative reasoning capabilities for agents that work with textual interactions and context memory. Earlier studies showed that agents, based on LLMs, can demonstrate social coherence as well as adapt to different dialogue environments.~\cite{gao2023s3, park2023generative, chuang2024simulating, rossetti2024social}. Nevertheless, current approaches are usually built upon synthetic interaction graphs or simplified agent models. Moreover, these approaches often lack rigorous validation in empirical grounding against real opinion dynamics.

Despite significant advances in classical and LLM-based approaches, a critical gap remains unaddressed. The classical approach (DeGroot, Friedkin--Johnsen, Hegselmann--Krause, and Deffuant--Weisbuch ) that models opinion change through numerical aggregation rules such as weighted averaging or confidence‑bounded updates provides structural clarity but falls short on providing any cognitive realism, whereas LLMs produce a lot of generative behavior but fail to validate their behaviors in real-world environments. 

To summarize, the research gaps we identified above include:

\begin{itemize}
\item Research Gap 1 (RG1): 
Classical opinion dynamics models provide strong mathematical foundations but offer limited cognitive and linguistic realism, as they do not account for the contextual, emotional, and semantic factors that shape opinion formation and change.

\item Research Gap 2 (RG2): 
LLM-based simulations rely on synthetic networks and simplified agent representations rather than real-world networks.

\item Research Gap 3 (RG3): 
Existing approaches provide limited evidence on whether LLM-based agents grounded in real user characteristics can reproduce the temporal evolution of real-world collective opinions.

\end{itemize}

\subsection{Research contributions}
To address the research gaps identified above, we make the following research contributions: 

\begin{itemize}
\item Research Contribution 1 (RC1): We propose a novel generative digital twin framework that clones empirical social network structures, including multidimensional user characteristics, with LLM-based agents to predict the trajectory of opinions, thereby introducing the cognitive and linguistic realism that classical models lack. 

\item Research Contribution 2 (RC2): We ground LLM-driven agents in an empirically cloned real-world Twitter network rather than a synthetic graph, instantiating each agent from attributes extracted from real user data, thereby providing the empirical grounding that prior LLM-based simulations lack.

\item Research Contribution 3 (RC3): We provide a validated predictive framework, demonstrating through evaluation on two real Twitter datasets and an extensive ablation study that agents grounded in real user characteristics reproduce the temporal evolution of individual-level and collective opinions, with agent traits being the essential contributor.

\end{itemize}

We organize the remainder of this paper as follows: Section~\ref{sec:related_work} reviews related work. Section~\ref{sec:problem} and Section~\ref{sec:Proposed_method} present the problem formulation and framework construction. Section~\ref{sec:experimental} and Section~\ref{sec:results} discuss the experimental setup and results, respectively. Section~\ref{sec:Discussion} discusses the results obtained, and Section~\ref{sec:limitations} presents limitations of our proposed framework. Finally, Section~\ref{sec:Conclusion} makes some concluding remarks.

\section{Related Work}
\label{sec:related_work}
The major theoretical background behind modeling opinion evolution in networked environments is built on dynamical systems. DeGroot model~\cite{degroot1974reaching} relies on opinion updating via weighted averaging, converging to a consensus state under proper connectivity conditions, whereas the Friedkin--Johnsen model~\cite{friedkin1990social} considers the individual agent's level of stubbornness. These past results have led to the development of bounded-confidence models: the Hegselmann--Krause model~\cite{rainer2002opinion} relies on restricting the interaction between agents based on confidence distance, leading to polarization and clustering. The Deffuant--Weisbuch model~\cite{deffuant2000mixing} uses pairwise stochastically driven opinion updates under the same restriction. Overall, these models are both mathematically tractable and provide an insightful analysis of consensus formation and social influence phenomena. Recent research efforts have been focusing on generalizing this theoretical background to more sophisticated opinion dynamics models that capture a broader range of social influence mechanisms and more complex patterns of opinion evolution~\cite{castellano2009statistical,proskurnikov2017tutorial}.

In this context, Liu et al.~\cite{liu2022probabilistic} proposed an adaptation of the DeGroot~\cite{degroot1974reaching} model to probabilistic linguistic data representing human opinions, but their approach remains limited to static linguistic consensus formation and cannot capture dynamic interaction processes in evolving social networks. Wu et al.~\cite{wu2022mixed} proposed hybrid DeGroot--Hegselmann--Krause models to analyze global aggregation and bounded‑confidence interaction, but their framework remains purely numerical and cannot represent the semantic content of opinions.  Conjeaud et al.~\cite{conjeaud2024degroot} developed a two‑layer DeGroot model with global information feedback showing how external signals (such as trending news feeds, press coverage, or election results) affect polarization, but the model still relies on numerical opinion updates and cannot represent the semantic content or reasoning behind agents’ opinions. Finally,  Nugent et al.~\cite{nugent2025opinion} developed a continuous age‑structured model reflecting the different response times of various population types (such as adolescents, working‑age adults, and elderly groups). At the same time, there is a clear limitation in the current approach: these approaches reduce the belief-update process to purely numerical rules, omitting the linguistically expressed reasoning behind real-life opinions~\cite{starnini2025opinion}.

The Agent-Based Modeling (ABM) approach builds upon this method by replacing analytical update rules with autonomous agents whose actions follow some behavioral rules~\cite{bonabeau2002agent,epstein2012generative,parsegov2016novel}. More recently, however, researchers have developed ABMs that address the issue of polarization via social media and how such platforms use the structural features of their networks, information diffusion, and algorithmic boosters to foster the spread of misinformation~\cite{cinelli2021echo,weismueller2024information}. Despite these developments, ABMs are still limited in that the behaviors of the agents follow pre-defined behavioral rules: they update their states by following manually defined mathematical functions rather than interpreting messages and their contexts.

The advent of LLMs has opened up an entirely new area of research. Park et al.~\cite{park2023generative} reveal that generative agents utilizing LLMs with memory and reflective capabilities exhibit emergent social behavior in sandboxed environments. Chuang et al.~\cite {chuang2024simulating} prove that networks of agents utilizing LLMs can be used to model opinion dynamics. Gao et al.~\cite{gao2023s3} present S3, the first LLM-enabled social network simulator, modeling emotion propagation and polarization through the analysis of actual world events. Mou et al.~\cite{mou2024unveiling} and Wang et al.~\cite{wang2025decoding} explore the phenomenon of echo chamber formation on scale-free, small-world, and random graphs with LLM-enabled agents and suggest nudging interventions to minimize polarization metrics. Despite such advancements, these agents exist in artificial environments devoid of empirical network reconstruction and rely mostly on qualitative comparisons for validation.

At a larger scale, AgentSociety exhibits the potential to synchronize over ten thousand agents in plausible social settings, according to Piao et al.~\cite{piao2025agentsociety}. 
Additionally, Park et al.~\cite{park2024generative} have found that interview-based agents can generate actual individual attitudes with a high level of precision. Recently, SocioVerse (Zhang et al.~\cite{zhang2025socioverse}) provides a world model in the context of social simulation based on a dataset of ten million actual user accounts obtained through social media, where alignment is attained along environmental, interaction, and behavioral aspects. While this innovation constitutes a significant step towards population-level reality, the proposed architecture focuses on behavioral trends at the aggregate level in politics, news, and economics rather than the temporal change of opinion measures in a simulated interaction setting. FDE-LLM~\cite{yao2025social} combines LLM-based role-playing and physically informed differential equations trained using Weibo data, but agents are still low-dimensional and function under an artificial social interaction lattice as opposed to a reconstructed social network. Two other works have developed platforms for simulating social digital twins without continuous, validated opinion-dynamics output. Rossetti et al.~\cite{rossetti2024social} reproduce user interactions, information diffusion, and network dynamics at the platform level, while G\"{u}rcan et al.~\cite{gurcan2025towards} build a synthetic-population simulator for scenario-based what-if analysis (e.g., youth school dropout). Neither work continuously simulates and validates quantitative opinion trajectories against empirical data. However, according to a recent systematic literature review by Larooij and T\"{o}rnberg~\cite{larooij2025validation}, there is one common problem among these models; namely, verification has remained an unresolved problem for these works due to the reliance on subjective evaluation of believability and the failure to achieve empirical validation at multiple scales.

\subsection{Summary of results and lessons learned from past related works}
We found that the current state‑of‑the‑art results achieved so far in the field of opinion dynamics simulation reveal three major gaps where novel solutions are needed. The first gap is that classical approaches (such as the DeGroot, Friedkin–Johnsen, Hegselmann--Krause, and Deffuant--Weisbuch models) simplify belief change into numerical update rules that ignore the semantic and emotional factors present in real social communication. The second gap is that existing LLM‑based simulations are typically built on synthetic networks rather than empirically reconstructed social graphs. The third gap is that current LLM‑driven frameworks lack validated predictive capabilities, as they rarely demonstrate that agent‑level and collective‑level opinion trajectories can be reproduced across multiple empirical scales. In this work, we address exactly this intersection by proposing a generative digital twin framework that integrates empirical network cloning, LLM‑driven transition operators, and a multi‑scale validation protocol.

\section{Problem Formulation}
\label{sec:problem}
\subsection{Digital Twin Background}
Digital twin technology provides a computational mirror of a real‑world system that mirrors its structure and behavior, enabling simulation and prediction without acting on the real system. Originally developed in engineering and manufacturing, the concept has recently been extended to social systems, demonstrating how artificial agents can replicate human interactions and collective behavior~\cite{rossetti2024social, gurcan2025towards}.  For instance, ~\cite{rossetti2024social} constructs an LLM‑driven digital twin of a social‑media platform to study emergent online dynamics, and ~\cite{gurcan2025towards} builds a data‑rich societal replica that enables interactive exploration of policy interventions. These efforts show that digital twins can serve as experimental sandboxes for understanding complex social processes. Motivated by this, we adopt the digital‑twin concept to replicate a real interaction network, evolve it autonomously under a generative model, and evaluate how closely the simulated trajectory matches the observed one. In the following sections, we formalize the empirical system, define its digital‑twin counterpart, and specify the objective linking the two.

\subsection{Empirical Social System}
Opinion dynamics unfold over a directed weighted graph $\mathcal{G} = (\mathcal{V}, \mathcal{E}, W)$, where $\mathcal{V} = \{1, \ldots, N\}$ is the set of agents, $\mathcal{E} \subseteq \mathcal{V} \times \mathcal{V}$ represents directed interactions, and $W = [w_{ji}]$ is the weighted adjacency matrix encoding empirical interaction strength. Each agent $i \in \mathcal{V}$ holds a time-dependent opinion state $o^t_i \in [-1, 1]$ (operationalized via sentiment polarity, see Section V-B-2) at discrete time $t \in \mathcal{T} = \{0, 1, \ldots, T\}$, where $-1$ denotes strong opposition, $+1$ denotes strong support, and $0$ represents neutrality. The global opinion configuration is the state vector $\mathbf{o}^t = (o^t_1, o^t_2, \ldots, o^t_N)^\top$. An unknown transition operator governs opinion evolution in the empirical system
\begin{equation}
\mathbf{o}^{t+1} = F^{\text{real}}(\mathcal{G},\, \mathbf{o}^t,\,
\theta),
\label{eq:real_transition}
\end{equation}

\noindent where $\theta$ encodes latent individual and social factors, such as persona, stubbornness, and emotional state that are not directly observable but are approximated through the empirically derived agent attribute vector $\mathbf{a}_i$ in the digital twin. The explicit form of $F^{\text{real}}$ is unknown because it reflects the linguistic and cognitive processes underlying real belief updates.

\subsection{Digital Twin System}
The digital twin is a computational system $\mathcal{G}^{DT} = (\mathcal{V}^{DT}, \mathcal{E}^{DT}, W^{DT})$ constructed to replicate the structural and agent-level propertiesof  of the empirical network. We enforce structural correspondence: each digital agent maps bijectively to its empirical counterpart, and we preserve the interaction topology exactly.

\begin{equation}
\mathcal{V}^{DT} \leftrightarrow \mathcal{V},\quad
\mathcal{E}^{DT} \leftrightarrow \mathcal{E},\quad
W^{DT} = W.
\label{eq:structural_correspondence}
\end{equation}

\noindent Each digital agent $i \in \mathcal{V}^{DT}$ is instantiated with an empirically derived attribute vector $\mathbf{a}^{DT}_i = \mathbf{a}_i$, preserving persona, structural position, behavioral patterns, and emotional features extracted during calibration. A generative transition operator governs opinion evolution in the twin.

\begin{equation}
\mathbf{o}^{t+1}_{DT} = F^{DT}(\mathcal{G}^{DT},\,
\mathbf{o}^t_{DT},\, \mathbf{A}^{DT}),
\label{eq:dt_transition}
\end{equation}

\noindent where $F^{DT}$ is implemented through LLM-driven generative agents and serves as a computational surrogate for the unknown $F^{\text{real}}$.

\subsection{Problem Objective}
The goal is to identify $F^{DT}$ such that the digital twin reproduces the empirical opinion trajectory over unseen time steps. 
Eq.~\eqref{eq:objective} expresses this requirement by stating that the opinion of each digital‑twin agent at time $t$ should closely match the real opinion observed at the same time during the validation period.

\begin{equation}
\mathbf{o}^t_{DT} \approx \mathbf{o}^t_{\text{real}},
\quad t \in \mathcal{T}_{\text{val}}.
\label{eq:objective}
\end{equation}

\noindent Approximation quality is operationalized across four complementary scales: individual-level accuracy (Mean Absolute Error (MAE)), distributional fidelity (Earth Mover's Distance (EMD)), polarization dynamics (Variance(var)), and structural alignment ($r$), which are formally defined in Section~\ref{subsec:Metrics}. This frames digital twin construction as a system identification problem: $F^{DT}$ is a surrogate dynamical system estimated from observed data, and its validity is assessed by how well it predicts opinion evolution outside the estimation window. $F^{DT}$ observes the opinion trajectory $\{\mathbf{o}^t_{\text{real}}\}$, the network $\mathcal{G}$, and agent-level content ($\theta$). $F^{DT}$ must capture the cognitive and social process ($F^{\text{real}}$).

\begin{figure*}[t]
\centering
\includegraphics[width=0.8\textwidth]{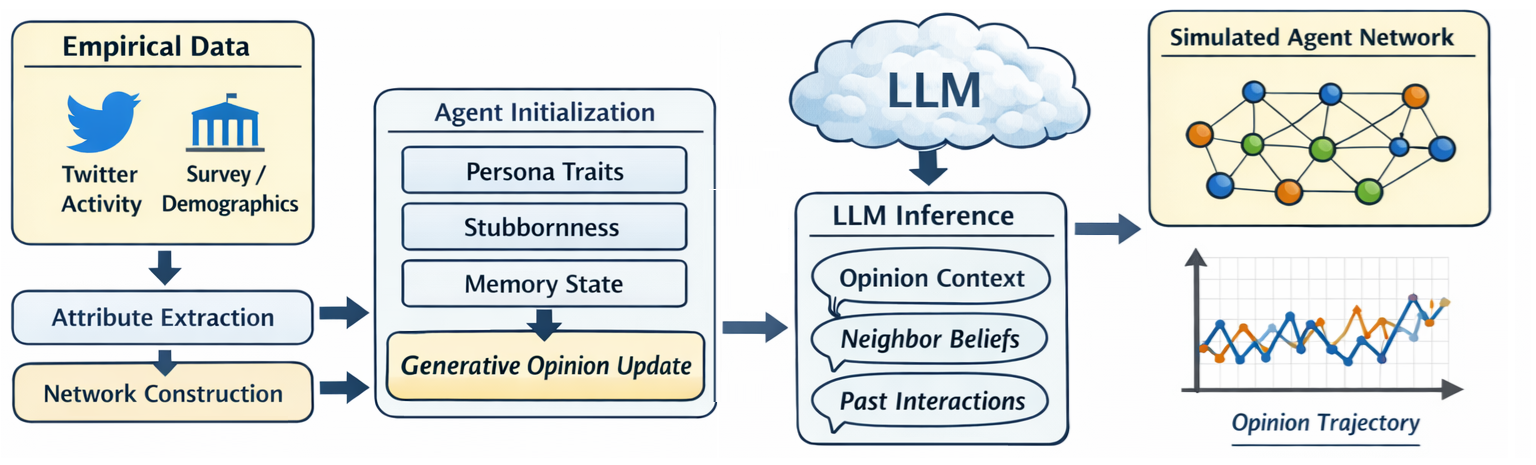}
\caption{Overview of the proposed generative digital twin framework. Empirical Twitter data is used to extract agent attributes and construct the interaction network. Each agent is initialized with persona traits, stubbornness, and a memory state, and opinions evolve through an LLM that conditions on the current opinion context, neighbor beliefs, and past interactions to produce an updated opinion trajectory.}
\label{fig:framework}
\end{figure*}

\section{Proposed Digital Twin Construction}
\label{sec:Proposed_method}
This section describes how the digital twin is constructed from empirical data. Fig.~\ref{fig:framework}thats the framework which transforms an empirical Twitter interaction network into a fully executable simulation. Three stages drive this process: the interaction topology is cloned exactly from real data, each agent is instantiated from empirically extracted attributes, and opinion evolution is driven by an LLM conditioned on memory and social exposure.  

\subsection{Structural Network Preservation}
We duplicate the interaction topology of the empirical network without modification. Each digital agent maps bijectively to its empirical counterpart, and we copy the adjacency matrix $W$ exactly. For each agent $i \in \mathcal{V}$, the neighbor set is $\mathcal{N}_i = \bigl\{ j \in \mathcal{V} \mid (j, i) \in 
\mathcal{E} \bigr\}$, and the corresponding digital agent preserves this set under structural correspondence: $\mathcal{N}_i^{DT} \leftrightarrow \mathcal{N}_i$. Every digital agent, therefore, operates under the same structural conditions as its real-world counterpart throughout the simulation.

\subsection{Agent Cloning via Empirical Attribute Instantiation}
We clone each empirical agent $i \in \mathcal{V}$ as a generative agent $\mathcal{A}_i^{DT}$ and assigned an attribute vector extracted during the calibration phase, $\mathbf{a}_i = \bigl(\mathbf{p}_i,\; \mathbf{s}_i,\; 
\mathbf{b}_i,\; \mathbf{c}_i,\; \lambda_i,\; \beta_i\bigr)$. where $\mathbf{p}_i$ encodes persona, $\mathbf{s}_i$ structural position, $\mathbf{b}_i$ behavioral activity, $\mathbf{c}_i$ emotional characteristics, $\lambda_i$ stubbornness, and $\beta_i$ influence. We describe each component below.

\subsubsection{Persona Features ($\mathbf{p}_i$)} Two components define agent persona: a unique identifier and a representation vector $\mathbf{p}_i \in \mathbb{R}^d$ encoding linguistic and communicative style derived from empirical tweet content. Persona features draw on Aristotle's classical rhetorical theory~\cite{demirdougen2010roots}, which identifies three foundational modes of persuasion:

\begin{itemize}
    \item Ethos (credibility): the degree to which an agent 
    establishes authority and trustworthiness through language.
    \item Pathos (emotional appeal): the extent to which the 
    agent evokes emotions, values, or affective responses.
    \item Logos (logical reasoning): the use of structured 
    arguments, evidence, and rational justification.
\end{itemize}

\subsubsection{Structural Features ($\mathbf{s}_i$)}
The structural position of each agent within the network is 
captured by $\mathbf{s}_i = (deg_i,\; PR_i,\; BC_i,\; CC_i)$, 
where:

\begin{itemize}
    \item $deg_i$: degree centrality reflects the number of direct connections the agent maintains.
    \item $PR_i$: PageRank score measures the global influence based on the importance of neighboring agents.
    \item $BC_i$: betweenness centrality quantifies the extent to which the agent acts as a structural broker.
    \item $CC_i$: closeness centrality captures how efficiently the agent reaches others in the network.
\end{itemize}

\subsubsection{Behavioral Metadata ($\mathbf{b}_i$)}
Behavioral activity is captured by $\mathbf{b}_i = (f_i,\; g_i,\; r_i,\; pr_i)$, where:
\begin{itemize}

\item $f_i$: follower count reflects the size of the agent's audience.
\item $g_i$: followee count represents subscriptions initiated by the agent.
\item $r_i$: interaction frequency measures the number of replies, comments, and reactions.
\item $pr_i$: posting rate captures the temporal intensity of content generation.
\end{itemize}

\subsubsection{Emotional Features ($\mathbf{c}_i$)}
Emotional features capture belief orientation and affective 
 Each agent carries a vector $\mathbf{c}_i = (o_i,\; \sigma_i,\; \mathbf{e}_i)$, where:

\begin{itemize}
\item $o_i$: opinion score quantifies the agent's position toward a target issue on a continuous numerical scale.
\item $\sigma_i$: sentiment polarity captures the overall positive or negative tendency in the agent's content.
\item $\mathbf{e}_i$: emotion vector encodes discrete emotional states such as anger, fear, and joy extracted from tweet text.
\end{itemize}

\subsubsection{Stubbornness and Influence}
Heterogeneous interaction behavior is captured by two empirically derived scalars. Stubbornness $\lambda_i \in [0,1]$ governs resistance to social influence; influence $\beta_i \in [0,1]$ measures the agent's capacity to shift the opinions of others.

\begin{itemize}

\item Stubbornness Estimation ($\lambda_i$): Stubbornness quantifies opinion persistence under social exposure. Let $m_i^t = \sum_{j \in \mathcal{N}_i} \tilde{w}_{ji}\, o_j^t$ denote the weighted neighbor mean at time $t$. The opinion update behavior is modeled as:

\begin{equation}
o_i^{t+1} = \lambda_i\, o_i^t + (1 - \lambda_i)\, m_i^t + 
\varepsilon_i^t,
\end{equation}

\noindent where $\varepsilon_i^t$ is a residual term. The parameter $\lambda_i$ is estimated by ordinary least-squares regression over calibration observations and clipped to $[0,1]$. Higher values indicate stronger belief persistence; lower values indicate greater openness to social influence.

\item Influence Score Estimation ($\beta_i$): The influence score $\beta_i$ measures the relative capacity of agent $i$ to propagate opinions across the network. Following the authors of \cite{2025digital}, it is derived from the fundamental matrix of the Friedkin-Johnsen \cite{friedkin1990social} model (an opinion dynamics model in which agents update their opinions by combining their initial opinions with the influence of their social neighbors) at equilibrium, which encodes how each agent's initial opinion propagates to all others. Formally, $\beta_i$ is the average impact of agent $i$'s initial opinion on the equilibrium opinions of the full agent population, normalized to $[0,1]$.

\end{itemize}

\subsection{Generative Transition Operator}
\label{subsec:generative_transition}

\subsubsection{Agent-Level Opinion Update}
For each digital agent $i \in \mathcal{V}^{DT}$, a generative transition function produces the opinion update at time $t$ by a generative transition function realized through Mistral-7B:

\begin{equation}
o_i^{t+1} = F_{\text{LLM}}\!\left(o_i^t,\; \mathcal{E}_i^t,\; 
\mathbf{a}_i,\; \mathcal{M}_i^t\right),
\label{eq:llm_update}
\end{equation}
\noindent where each input is defined as follows:

\begin{itemize}
\item $o_i^t$: the agent's opinion at time $t$, serving as its internal belief state before updating and conditioning the entire generative transition.

\item $\mathcal{E}_i^t$: social exposure at time $t$. For each neighbor $j \in \mathcal{N}_i^{DT}$, the weighted opinion $\tilde{w}_{ji} o_j^t$ and the neighbor's most recent post $x_j$ are combined into a single formatted string. The textual component $x_j$ is fixed at its calibration-period value for each neighbor and is not refreshed during validation; only the neighbor's opinion signal $\tilde{w}_{ji} o_j^t$ is updated at each validation step via the simulation. This ensures that no future empirical content enters the simulation after the calibration boundary.

\item $\mathbf{a}_i$: the empirically derived attribute vector encoding persona traits, structural position, behavioral patterns, emotional characteristics, stubbornness, and influence held fixed across all timesteps.

\item $\mathcal{M}_i^t$: the agent's memory state. At each timestep, it is updated by appending the new opinion $o_i^{t+1}$ and the full LLM-generated reasoning text. No window limit is applied: the complete history from the calibration boundary is retained in subsequent prompts, enabling longitudinal coherence at the cost of increasing prompt length over time, a trade-off discussed in Section~\ref{sec:limitations}.

\end{itemize}

\subsubsection{Prompt Structure}
\begin{figure}[t]
\centering
\includegraphics[width=0.7\columnwidth]{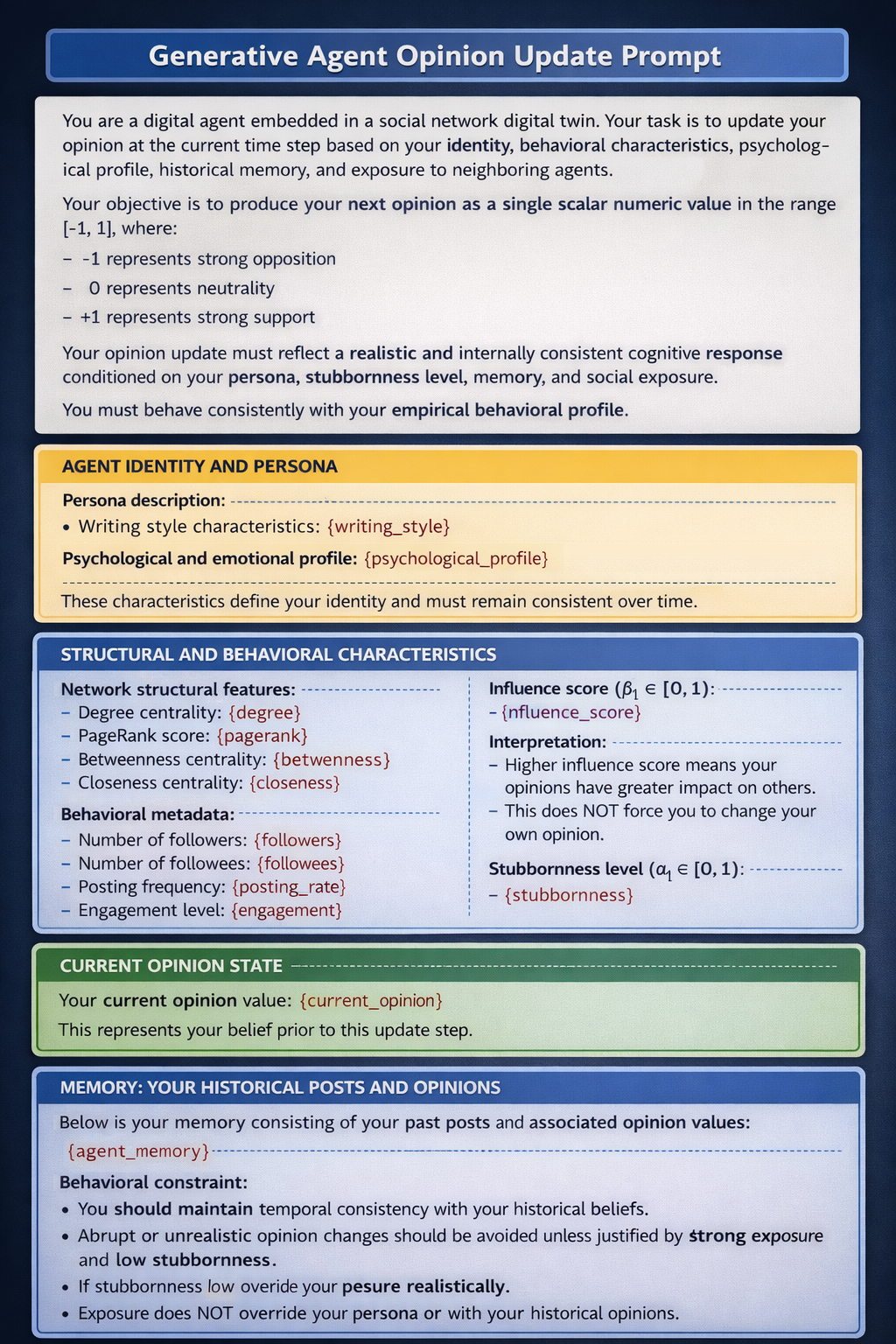}
\caption{Structured prompt template used to generate opinion updates for digital agents.}
\label{fig:llm_prompt}
\end{figure}

$F_{\text{LLM}}$ conditions the model on the agent's full internal state and social context through a structured prompt (Fig.~\ref{fig:llm_prompt}). The ordering is deliberate. Agent identity and persona appear first, anchoring the model's behavioral profile before any external signal is introduced. Structural and behavioral features follow, establishing how active and influential the agent is within the network. The current opinion state appears immediately before the memory block, so the model processes its present position and historical trajectory together. Neighbor exposure appears last, reflecting the natural cognitive order: an agent consolidates its own prior beliefs before processing incoming social signals. The prompt instructs the model to return an updated opinion consistent with the agent's persona, prior beliefs, and observed interactions. A heuristic extraction rule parses the response that retrieves the first numeric value in the output and projects it onto $[-1, 1]$. This is simple and consistent with all agents

\subsection{Simulation Procedure} 
The digital twin evolves autonomously over the validation horizon through synchronized multi-agent updates. Algorithm~\ref{alg:dt_rollout} summarizes the full procedure. At the calibratio.n boundary, every agent is initialized with its empirical opinion $o_i^{T_{cal}}$ and attribute vector $\mathbf{a}_i$. At each step $t \in \mathcal{T}_{val}$, agents observe neighbor states under the preserved topology and compute their next opinion via Eq.~\eqref{eq:llm_update}. The resulting opinion and reasoning text are appended to $\mathcal{M}_i^t$ to maintain longitudinal coherence. Executing this rule synchronously across all agents yields the global opinion vector $\mathbf{o}_{DT}^{t+1}$. Repeating over $\mathcal{T}_{val}$ produces the simulated trajectory

\begin{equation}
\mathcal{O}_{DT} = \bigl\{\mathbf{o}_{DT}^t\bigr\}_{t=T_{cal}}^{T},
\end{equation}

\noindent which is compared against the empirical trajectory $\mathcal{O}_{real}$ to assess predictive fidelity across all four evaluation scales defined in Section~\ref{subsec:Metrics}.

\begin{algorithm}[t]
\caption{Digital Twin Simulation}
\label{alg:dt_rollout}
\begin{algorithmic}[1]
\State \textbf{Input:} Calibration network $G^{\mathrm{DT}}=(V,E,W)$; attributes $A^{\mathrm{DT}}=\{\mathbf{a}_i\}$; boundary state $\mathbf{o}^{T_{\mathrm{cal}}}_{\mathrm{real}}$; horizon $T$
\State \textbf{Initialize agents:} for each $i\in V$, set $\mathcal{A}_i^{\mathrm{DT}}\leftarrow (i,\mathbf{a}_i,o_i^{T_{\mathrm{cal}}},\mathcal{M}_i^{T_{\mathrm{cal}}})$
\State \textbf{Preserve network:} set $(E^{\mathrm{DT}},W^{\mathrm{DT}})\leftarrow (E,W)$
\State \textbf{Initialize state:} $\mathbf{o}^{T_{\mathrm{cal}}}_{\mathrm{DT}} \leftarrow \mathbf{o}^{T_{\mathrm{cal}}}_{\mathrm{real}}$
\State \textbf{Record:} $\mathcal{O}_{\mathrm{DT}} \leftarrow \{\mathbf{o}^{T_{\mathrm{cal}}}_{\mathrm{DT}}\}$

\For{$t=T_{\mathrm{cal}},\dots,T-1$} \Comment{Iterate over time}
    \ForAll{$i\in V$} \Comment{Compute next opinions (synchronous)}
        \State Compute normalized weights $\tilde{w}_{ji}$ for $j\in N_i$
        \State Compute exposure $\mathcal{E}^t_i$: for each $j \in \mathcal{N}_i$, format $(\tilde{w}_{ji} o^t_j,\; x_j)$ as a single string ($x_j$ fixed from calibration, $o^t_j$ updated via simulation); concatenate across neighbors

        \State Compute epistemic input $H_i^t \leftarrow \beta_i E_i^t$
        \State Generate $o_i^{t+1} \leftarrow F_{\mathrm{LLM}}(o_i^t,H_i^t,\mathbf{a}_i,\mathcal{M}_i^t)$
        \State Project $o_i^{t+1} \leftarrow \Pi_{[-1,1]}(o_i^{t+1})$
        \State $\mathcal{M}^{t+1}_i \leftarrow \mathcal{M}^t_i \;\|\; (o^{t+1}_i,\; \text{reasoning}^{t+1}_i)$ \hfill $\triangleright$ Append opinion and LLM reasoning text
        
    \EndFor
    \State Set $\mathbf{o}^{t+1}_{\mathrm{DT}} \leftarrow (o_1^{t+1},\dots,o_N^{t+1})^\top$ \Comment{Update opinions}
    \State Append $\mathbf{o}^{t+1}_{\mathrm{DT}}$ to $\mathcal{O}_{\mathrm{DT}}$ \Comment{Record trajectory}
\EndFor

\State \textbf{Output:} Digital twin trajectory $\mathcal{O}_{\mathrm{DT}}$
\end{algorithmic}
\end{algorithm}

\section{Experimental Setup}
\label{sec:experimental}

This section describes the empirical setup used to evaluate our proposed framework. Next, we describe the datasets, preprocessing procedures, experimental protocol, model implementations, and evaluation practices that we used in our study.  

\subsection{Datasets and Preprocessing}
\label{subsec:datasets}
\subsubsection{Datasets}
We used two publicly available Twitter datasets, each capturing real-world opinion dynamics in a distinct socio-political context. Both are collected from Kaggle \cite{chakraborty_covid19_twitter, manchunhui_us_election_2020} and consist of tweet content alongside user-level metadata, including follower counts, engagement statistics, and timestamps.

The COVID-19 Twitter dataset \cite{chakraborty_covid19_twitter} contains more than 1.1 million tweets related to the COVID-19 pandemic across three temporal periods spanning approximately one month each, merged into a unified observation window. 

The U.S.\ Election 2020 Twitter dataset \cite{manchunhui_us_election_2020} contains over 1.7 million tweets collected using the \#Trump and \#Biden hashtags during the 2020 U.S.\ presidential election and includes user location, likes, and retweet counts.

We processed all tweets through a standardized normalization pipeline: raw text was lowercased; URLs, user mentions, and HTML artifacts were removed via regular expressions; hashtag symbols were stripped while preserving lexical tokens; punctuation, numerical tokens, and non-ASCII characters were filtered; and emojis were removed unless they contributed to sentiment interpretation. We tokenized the cleaned text using the RoBERTa tokenizer. We removed stopwords using the NLTK English stopwords list\cite{bird2009natural}. We removed duplicate tweets and retweets to prevent aggregation bias. This pipeline preserves the linguistic signals required for sentiment, persuasion, and emotional feature extraction.

\subsubsection{Network Construction and Agent Filtering}
\label{subsubsec:network}
We construct the empirical social network $\mathcal{G} = (\mathcal{V}, \mathcal{E}, W)$ from mention-based interactions. We establish a directed edge $(j, i) \in \mathcal{E}$ when user $j$ mentions user $i$ in a tweet, and the interaction weight $w_{ji}$ is defined as the mention frequency from $j$ to $i$ over the full observation window. We identify the community structure using the Leiden algorithm~\cite{traag2019louvain}, which outperforms Louvain’s algorithm \cite{blondel2008fast} on modularity optimization for directed graphs by guaranteeing well-connected communities. We retain the largest connected component to ensure structural coherence and sufficient interaction density for simulation. The COVID-19 dataset yields approximately 700 agents (880 calibration edges, 377 validation edges); the election dataset yields approximately 500 agents (677 calibration edges, 290 validation edges).

\subsection{Agent Attribute Extraction}
\label{subsec:attributes}
We extracted user-level attributes from the calibration phase to construct the empirically grounded agent attribute vector $\mathbf{a}_i$.

\subsubsection{Persuasion Strategies} 
We detect rhetorical persona features using a fine-tuned multilingual transformer trained for SemEval 2024 Task 4~\cite{chatterjee2024multilingual}. Post-level persuasion labels are mapped to Aristotle's rhetorical categories \cite{braet1992ethos}: ethos, pathos, and logos. We aggregate it per user to obtain normalized distributions. Table~\ref{tab:posts-table-election-vax} presents illustrative examples of each category.

\begin{table*}[ht]
\centering
\caption{Examples of persuasion types (ethos, logos, pathos).}
\label{tab:posts-table-election-vax}
\begin{tabular}{|c|p{5.5cm}|p{5.5cm}|}
\hline
\textbf{Persuasive type} & \textbf{Example post} & \textbf{Explanation} \\ \hline
{Ethos} & \textit{“As a former U.S. election official, I can confirm that the voting process is secure and closely monitored to ensure accuracy.”} & Builds credibility by referencing professional experience and authority in election administration. \\ \hline
{Logos} & \textit{“CDC data shows that vaccinated individuals are significantly less likely to be hospitalized. The latest report indicates over 85\% reduction in severe cases.”} & Uses statistical evidence and references to official data to logically support the argument for vaccination. \\ \hline
{Pathos} & \textit{“Every vote represents a voice that deserves to be heard. Don’t let your community’s future be decided without you—show up and vote!”} & Appeals to emotions such as civic pride and responsibility to motivate participation in the election process. \\ \hline
\end{tabular}
\end{table*}

\subsubsection{Emotional Attributes and Opinion Construction} 
We performed sentiment analysis using a pre-trained RoBERTa model~\cite{liu2019roberta} to produce probability distributions over negative, neutral, and positive classes. These are aggregated to compute daily user opinions $o_i^t \in [-1,1]$, forming the empirical trajectory $\{\mathbf{o}_{\mathrm{real}}^t \}_{t \in \mathcal{T}}$. We extracted emotion intensities using the DistilRoBERTa model~\cite{hartmann2022emotionenglish}, yielding per-user scores for anger, fear, joy, and related discrete states.

\subsubsection{Stubbornness and Influence Estimation} Stubbornness $\lambda_i$ is estimated via per-user ordinary least-squares regression,$o_i^{t+1} = \lambda_i\, o_i^t + (1 - \lambda_i)\, m_i^t +
\varepsilon_i^t, \quad
m_i^t = \sum_{j \in \mathcal{N}_i} \tilde{w}_{ji}\, o_j^t$.= where $\varepsilon_i^t$ is a residual term. We clipped the estimated $\lambda_i$ to $[0,1]$. 
 
We derived influence scores $\beta_i$ from the Friedkin--Johnsen fundamental matrix $\mathbf{H} = (\mathbf{I} - \mathbf{A}\mathbf{W})^{-1} (\mathbf{I} - \mathbf{A})$, which encodes how each agent's initial opinion propagates to all others at equilibrium. Each $\beta_i$ is defined as the average impact of agent $i$'s initial opinion on all equilibrium opinions, normalized to $[0,1]$.
 
\subsection{Baseline Models}
\label{subsec:baselines}
Four classical opinion dynamics models serve as baselines,
evaluated on the same empirical network $\mathcal{G}^{DT}=(\mathcal{V}, \mathcal{E}, W)$ and validation horizon
$\mathcal{T}_{\mathrm{val}}$. Each model is described by its equation (Eq.~\eqref{eq:dg}, Eq.~\eqref{eq:fj}, Eq.~\eqref{eq:hk}, Eq.~\eqref{eq:DW} ) and the parameter range.

The DeGroot (DG) model~\cite{degroot1974reaching} updates opinions as weighted neighbor averages:
\begin{equation}
o_i^{t+1} = \sum_{j \in \mathcal{N}_i \cup \{i\}}
w_{ji}\, o_j^t, \quad \sum_j w_{ji} = 1.
\label{eq:dg}
\end{equation}
It has no tunable hyperparameters, and it is executed directly using calibrated interaction weights.

The Friedkin--Johnsen (FJ) model~\cite{friedkin1990social} balances social influence against intrinsic opinion persistence, where $\alpha_i \in [0,1]$ is susceptibility to social influence and $s_i$ is the intrinsic opinion from calibration data. Since we derive each agent's influence score $\beta_i$ from the Friedkin--Johnsen fundamental matrix (Section~\ref{sec:Proposed_method}), we note that its susceptibility parameter $\alpha_i$ is distinct from the stubbornness coefficient $\lambda_i$ used in our digital twin. The susceptibility grid is $\alpha \in \{0.0, 0.2, 0.4, 0.6, 0.8, 1.0\}$. Eq.~\eqref{eq:fj} defines the Friedkin--Johnsen update, in which each agent blends the weighted average of its neighbors' opinions with its own intrinsic opinion:

\begin{equation}
o_i^{t+1} = \alpha_i \sum_{j \in \mathcal{N}_i}
w_{ji}\, o_j^t + (1 - \alpha_i)\, s_i.
\label{eq:fj}
\end{equation}

where $\alpha_i \in [0,1]$ is agent $i$'s susceptibility to social influence, $w_{ji}$ is the interaction weight from neighbor $j$, $o^t_j$ is neighbor $j$'s opinion at time $t$, and $s_i$ is the intrinsic (anchor) opinion fixed from calibration data.

The Hegselmann--Krause (HK) model~\cite{rainer2002opinion} restricts influence to agents within a confidence threshold. Eq.~\eqref{eq:hk} defines the Hegselmann--Krause update, where each agent averages only the opinions of neighbors that lie within a confidence threshold of its own:

\begin{equation}
o_i^{t+1} = \frac{1}{|\mathcal{C}_i^t|}
\sum_{j \in \mathcal{C}_i^t} o_j^t, \quad
\mathcal{C}_i^t = \bigl\{ j \in \mathcal{N}_i \cup \{i\}
\mid |o_j^t - o_i^t| \le \epsilon \bigr\}.
\label{eq:hk}
\end{equation}
with $\epsilon \in \{0.0, 0.2, 0.4, 0.6, 0.8, 1.0\}$.

where $C^t_i$ is the confidence set of agent $i$ at time $t$, $\epsilon$ is the confidence threshold (the maximum opinion distance within which two agents influence each other), and $|C^t_i|$ is the number of agents in that set. Neighbors whose opinions differ by more than $\epsilon$ are excluded from the update.

The Deffuant--Weisbuch (DW) model~\cite{deffuant2000mixing} applies pairwise stochastic adjustments under the same bounded-confidence constraint. Eq.~\eqref{eq:dw} defines the Deffuant--Weisbuch update, a pairwise stochastic rule under the same bounded-confidence constraint; we include it as a baseline because it represents the pairwise (rather than simultaneous) family of bounded-confidence models:
 
\begin{equation}
\text{if } |o_i^t - o_j^t| \le \epsilon:\quad
o_i^{t+1} = o_i^t + \mu(o_j^t - o_i^t).
\label{eq:DW}
\end{equation}

where $\mu \in (0, 0.5]$ is the adjustment rate controlling how far two interacting agents move toward each other, and $\epsilon \in \{0.0, 0.2, 0.4, 0.6, 0.8, 1.0\}$ is the confidence threshold as in Eq.~\eqref{eq:hk}. An update occurs only when $|o^t_i - o^t_j| \le \epsilon$.

\subsection{Evaluation Metrics}
\label{subsec:Metrics}
We evaluated the simulated opinion trajectory $\mathcal{O}_{\mathrm{DT}}$ against the empirical trajectory $\mathcal{O}_{\mathrm{real}}$ using four complementary metrics that capture different levels of agreement. For each metric $\mathcal{M}$, the discrepancy is defined as $\Delta\mathcal{M} = \left| \mathcal{M}(\mathbf{o}_{\mathrm{DT}}) - \mathcal{M}(\mathbf{o}_{\mathrm{real}}) \right|$, where metric values are computed from the simulated and empirical opinion vectors. $\Delta\mathcal{M} = 0$ indicates perfect reproduction of the empirical dynamics whereas larger values indicate increasing divergence.

\begin{itemize}

   \item {\bf Mean Absolute Error (MAE)}: This metric is used to assess whether the digital twin accurately reproduces individual-level opinion dynamics by quantifying how closely the simulated opinions match empirical opinions at the agent level:

    \begin{equation}
    \mathrm{MAE}(t) = \frac{1}{|\mathcal{V}|} \sum_{i \in \mathcal{V}}
    |o_{i,\mathrm{real}}^t - o_{i,\mathrm{DT}}^t|,
    \end{equation}
    where $o^t_{i,real}$ and $o^t_{i,DT}$ are the empirical and simulated opinions of agent $i$ at time $t$, respectively, and $|V|$ is the number of agents.

    \item {\bf Earth Mover’s Distance ($EMD$)}: This metric is used to evaluate whether the digital twin correctly captures the global opinion distribution by measuring distributional mismatch between simulated and empirical opinion states:

    \begin{equation}
    \mathrm{EMD}^t = W_1\!\left( \mu^t_{\mathrm{DT}}, \mu^t_{\mathrm{real}} \right).
    \end{equation}
    where $W_1$ denotes the first Wasserstein (Earth Mover's) distance, and $\mu^t_{DT}$ and $\mu^t_{real}$ are the opinion distributions of the simulated and empirical networks at time $t$, respectively.
    
    \item {\bf Variance ($Var$)}: It captures the level of opinion dispersion in the network. This metric measures whether the simulated opinion can reproduce polarization phenomena.

    \begin{equation}
    \mathrm{Var}(\mathbf{o}^t) = \frac{1}{|\mathcal{V}|}
    \sum_{i \in \mathcal{V}} (o_i^t - \hat{o}^t)^2,
    \end{equation}
    where $o^t_i$ is the opinion of agent $i$ at time $t$, $\hat{o}^t$ is the mean opinion across all agents at time $t$, and $|V|$ is the number of agents.
    
    \item {\bf Opinion homophily ($r$)}: It measures the tendency of connected nodes in sharing similar opinion values, computed as the Pearson correlation between the opinions of neighboring nodes.

    \begin{equation}
    r(t) = \frac{\displaystyle\sum_{(i,j) \in \mathcal{E}}
    (o_i^t - \bar{o}^t)(o_j^t - \bar{o}^t)}
    {\displaystyle\sum_{(i,j) \in \mathcal{E}}
    (o_i^t - \bar{o}^t)^2},
    \end{equation}
    
    where $o^t_i$ and $o^t_j$ are the opinions of connected agents $i$ and $j$, $\bar{o}^t$ is the mean opinion at time $t$, and $E$ is the edge set. Here $r(t)$ is the Pearson correlation between the opinions of neighboring nodes.
    
\end{itemize}

\subsection{Implementation Details}
\label{subsec:implementation} 
\subsubsection{Digital Twin (Mistral-7B)} 
We implemented the generative transition operator $F_{\mathrm{LLM}}$ using Mistral-7B~\cite{jiang2023mistral} via the LangChain framework\cite{langchain}. We selected Mistral-7B because of its strong performance and its low inference cost relative to proprietary alternatives. Each agent operates as an autonomous generative entity with memory, attribute conditioning, and exposure inputs. The sensitivity analysis focuses on the temperature parameter $\tau$, which controls output stochasticity, across six values: $\text{Low: } \{0.0,0.2\}, \text{Medium: } \{0.4,0.6\}, \text{High: } \{0.8,1.0\}$ . We kept all other inference parameters and prompt templates fixed across runs to isolate the effect of generative variability. We averaged each of the four evaluation metrics over 15 independent runs per configuration.

\subsubsection{Baseline Hyperparameter Configuration,}
We evaluated each tunable baseline across six configurations. For Friedkin-Johnsen, the susceptibility parameter varies over $\alpha \in
\{0.0, 0.2, 0.4, 0.6, 0.8, 1.0\}$. For Hegselmann--Krause, and Deffuant--Weisbuch models, the confidence threshold varies over $\epsilon \in \{0.0, 0.2, 0.4,
0.6, 0.8, 1.0\}$. DG has no tunable parameters and is run once using calibrated interaction weights. This yields 19 configurations in total across all baselines.

\subsection{Experimental Protocol}
\label{subsec:protocol}

\subsubsection{Calibration, Validation, and Stochasticity} 
We divided the observation window $\mathcal{T} = \{0, \ldots, T\}$ into a calibration phase $\mathcal{T}_{\mathrm{cal}} = \{0, \ldots, T_{\mathrm{cal}}\}$ covering 70\% of timesteps and a validation phase $\mathcal{T}_{\mathrm{val}} = \{T_{\mathrm{cal}} + 1, \ldots, T\}$ covering the remaining 30\%. Agents active in both phases are retained to form a stable simulation population $\mathcal{V}^*$. The digital twin is initialized at the calibration boundary using the empirical opinion state $\mathbf{o}_{\mathrm{real}}^{T_{\mathrm{cal}}}$ and evolves autonomously over $\mathcal{T}_{\mathrm{val}}$ without further access to observed data. This protocol ensures that evaluation is based on out-of-sample prediction rather than in-sample fit. 

The digital twin is inherently stochastic due to LLM behaviors. To quantify this variability, we conducted 15 independent simulation runs per configuration using different random seeds. We computed the various metrics per run and aggregated them. We reported results as mean $\pm$ 95\% confidence interval, computed via normal approximation: $\bar{x} \pm 1.96\, s/\!\sqrt{n}$, where $n = 15$ and $s$ is the empirical standard deviation across runs. As the classical baselines are deterministic (fixed networks, initial opinions), and parameters produce identical trajectories across runs, so single-run evaluation was sufficient.

\subsubsection{Model Selection Protocol} 
For each model, we selected the best-performing hyperparameter configuration after evaluating all four metrics: MAE, $\Delta$EMD, $\Delta$Var, and $\Delta r$ individually over the validation horizon. We selected the configuration that achieved the lowest discrepancy across the greatest number of metrics.  

\subsubsection{Ablation Study Design} 
We evaluated three ablation variants, each removing one core component of the digital twin. These three components include agent attributes, memory, and social exposure, and they are the only inputs to $F_{\mathrm{LLM}}$ beyond the current opinion state, making them the natural candidates for ablation. Removing the network topology or the LLM itself would produce a fundamentally d,ifferent model class rather than a controlled ablation of the proposed framework.

\paragraph{Ablation 1. Removal of Agent Attributes}
The attribute vector $\mathbf{a}_i$ is excluded from the prompt. $F_{\mathrm{LLM}}$ is conditioned only on the prior opinion and external inputs, producing homogeneous agent behavior. This isolates the contribution of persona, structural, behavioral, and emotional heterogeneity to predictive fidelity. 

\paragraph{Ablation 2. Removal of memory $\mathcal{M}_i^t$} 
The memory module is removed. The transition operator is conditioned solely on the current state and instantaneous social context, preventing agents from accessing historical information about their past opinions and prior interactions with neighbors. This isolates the contribution of longitudinal belief coherence. 

\paragraph{Ablation 3. Removal of social exposure $\mathcal{E}_i^t$} 
In this experiment, neighbor opinions and the text of their posts are excluded. $F_{\mathrm{LLM}}$ is conditioned only on internal agent state, transforming the system into a set of independent generative processes. This isolates the contribution of network-coupled interaction to collective opinion formation.

\section{Experimental Results} 
\label{sec:results} 
This section evaluates the predictive fidelity of Mistral-7B as a generative opinion simulation engine across four complementary scales. We organize the results as follows: overall performance against all baselines (Section~\ref{subsec:overall}), temporal trajectory analysis (Section~\ref{subsec:temporal}), ablation study (Section~\ref{subsec:ablation}), and sensitivity analysis (Section~\ref{subsec:sensitivity}).

\subsection{Overall Predictive Performance} \label{subsec:overall} 
\begin{figure}[t]
\centering
\includegraphics[width=\columnwidth]{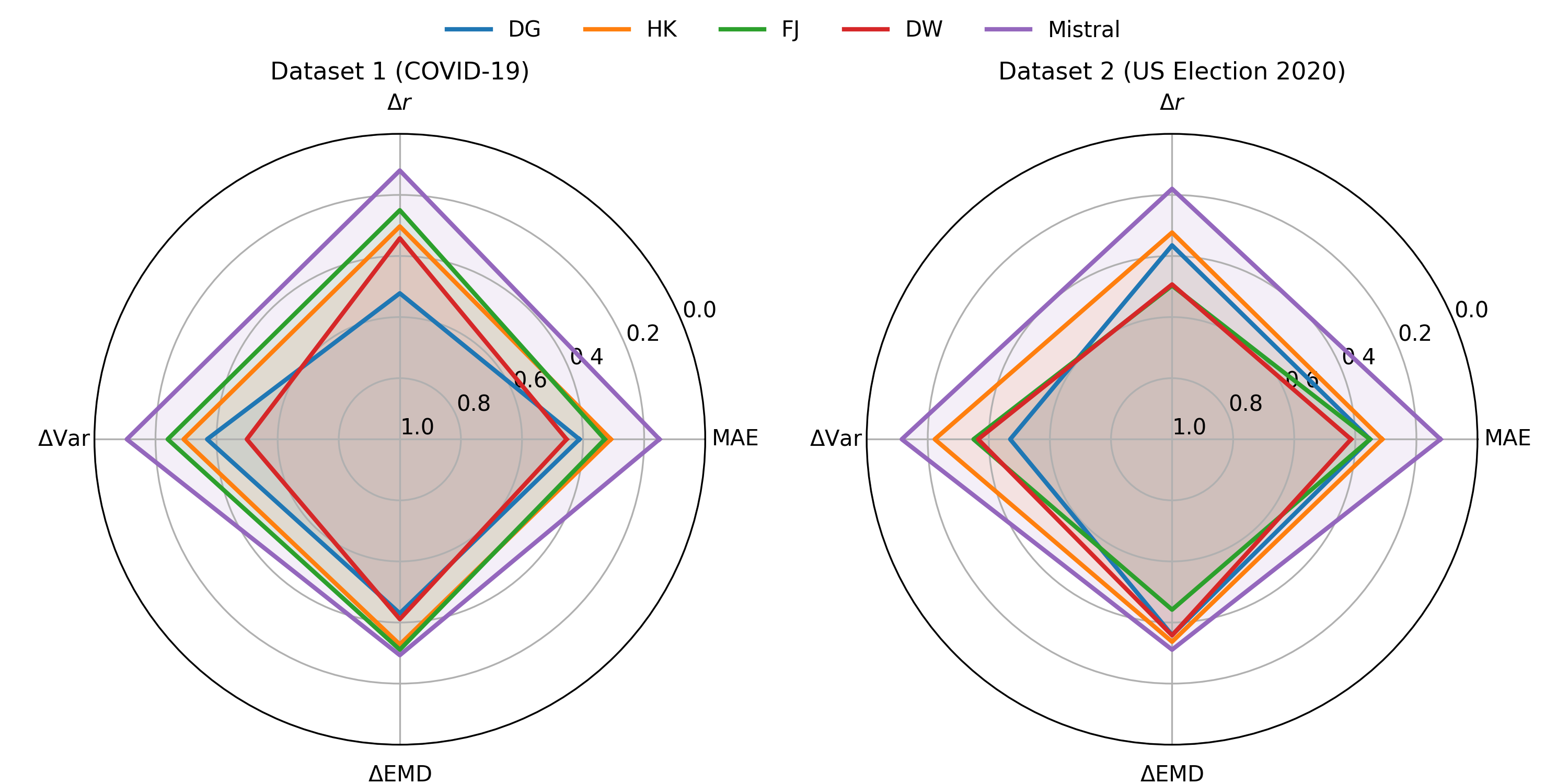}
\caption{Model performance comparison across datasets.DG: DeGroot; FJ: Friedkin--Johnsen; HK: Hegselmann--Krause; DW: Deffuant--Weisbuch; Param: model parameter; MAE: mean absolute error; $r$: homophily; Var: variance; EMD: Earth Mover's Distance.}
\label{fig:comparison}
\end{figure}

Table~\ref{tab:overall_fidelity} presents the full results; Fig.~\ref{fig:comparison} provides a visual comparison across datasets. Mistral-7B achieves the lowest mean discrepancy across all four evaluation metrics on both datasets, with no exception. The strongest classical baseline varies by metric: Hegselmann--Krause achieves the lowest MAE among classical models on both datasets, while Friedkin--Johnsen achieves the lowest $\Delta r$ on the COVID-19 dataset. Mistral-7B outperforms all four baselines regardless of this variation.

\subsubsection{Individual-level accuracy.}
Mistral-7B achieves MAE = 0.150 $\pm$ 0.02 on the COVID-19 dataset and MAE = 0.121 $\pm$ 0.02 on the election dataset, reducing the individual prediction error by more than 50\% relative to the best-performing classical model on each dataset (Hegselmann--Krause: 0.309 and 0.312, respectively). Classical models cluster between 0.309 and 0.454 on COVID-19, and between 0.312 and 0.414 on the election dataset, reflecting the ceiling imposed by fixed numerical update rules that cannot adapt to the linguistic content of social interactions.
These results demonstrate that the best-performing classical baseline's value (Hegselmann--Krause: 0.309 and 0.312) falls outside Mistral-7B's 95\% confidence interval on both datasets; the observed improvement is unlikely to be attributable to run-to-run stochastic variation alone.

\begin{table}[t]
\centering
\caption{Overall digital twin fidelity (validation). DG: DeGroot; FJ: Friedkin--Johnsen; HK: Hegselmann--Krause; DW: Deffuant--Weisbuch; Param: model parameter; MAE: mean absolute error; $r$: homophily; Var: variance; EMD: Earth mover's distance.}
\label{tab:overall_fidelity}
\footnotesize
\setlength{\tabcolsep}{3.5pt}
\renewcommand{\arraystretch}{1.05}
\resizebox{\columnwidth}{!}{
\begin{tabular}{lccccc}
\hline
Model & Param & MAE & $\Delta r$ & $\Delta$Var & $\Delta$EMD \\
\hline
\multicolumn{6}{c}{\textbf{Dataset 1 (COVID-19)}} \\
\hline
DG   & -- & 0.412 & 0.522 & 0.369 & 0.430 \\
HK   & $\varepsilon=0.2$ & 0.309 & 0.303 & 0.292 & 0.328 \\
FJ   & $\lambda=0.4$ & 0.329 & 0.250 & 0.240 & 0.311 \\
DW   & $\varepsilon=0.4$ & 0.454 & 0.342 & 0.499 & 0.411 \\
Mistral  & $\tau=0.8$ & $0.150$±0.02 & $0.120$±0.05 & $0.106$±0.04& $0.293$±0.10 \\
\hline
\multicolumn{6}{c}{\textbf{Dataset 2 (US Election 2020)}} \\
\hline
DG   & -- & 0.352 & 0.365 & 0.470 & 0.359 \\
HK   & $\varepsilon=0.2$ & 0.312 & 0.323 & 0.223 & 0.337 \\
FJ   & $\lambda=0.6$ & 0.352 & 0.497 & 0.350 & 0.442 \\
DW   & $\varepsilon=0.4$ & 0.414 & 0.493 & 0.365 & 0.358 \\
Mistral  & $\tau=0.8$ & $0.121$±0.02 & $0.180$±0.07& $0.115$±0.01 &  $0.311$±0.15 \\
\hline
\end{tabular}
}
\end{table}

\subsubsection{Polarization dynamics.} 
Mistral-7B achieves $\Delta\mathrm{Var} = 0.106 \pm 0.04$ and $0.115 \pm 0.01$ on the two datasets, representing reductions of 55.8\% and 48.4\% over the best classical baselines (Friedkin--Johnsen: 0.240 on COVID-19; Hegselmann--Krause: 0.223 on the election dataset). Classical models exhibit discrepancies ranging from 0.192 to 0.499, reflecting their tendency to either collapse toward artificial consensus or rigidly cluster under bounded-confidence constraints. As with MAE, the best classical baseline's value falls outside Mistral-7B's confidence interval on both datasets, indicating a statistically robust improvement.

\subsubsection{Structural alignment.} 
Mistral-7B achieves $\Delta r = 0.120 \pm 0.05$ and $0.180 \pm 0.07$ on the two datasets(COVID-19 and US election 2020), representing reductions of 52.0\% and 44.3\% over the best classical baselines (Friedkin--Johnsen: 0.250 on COVID-19; Hegselmann--Krause: 0.323 on the election dataset). Classical baselines range from 0.250 to 0.522, confirming that Mistral-7B preserves network-level opinion alignment between connected agents with substantially greater fidelity than any classical model, while the remaining discrepancy indicates room for further improvement in structural reproduction. As with the previous metrics, the corresponding classical baseline values (Friedkin--Johnsen: 0.250; Hegselmann--Krause: 0.323) fall outside Mistral-7B's confidence interval, supporting a statistically robust improvement.

\subsubsection{Distributional fidelity.} 
Mistral-7B achieves $\Delta\mathrm{EMD} = 0.293 \pm 0.10$ on COVID-19 and $0.311 \pm 0.15$ on the election dataset. These mean values remain below all classical baselines on COVID-19 (best: Friedkin--Johnsen 0.311) and are competitive on the election dataset (best: Hegselmann--Krause 0.337). However, the confidence intervals are notably wide, particularly on the election dataset, where $\pm 0.15$ reflects substantial run-to-run variability. This indicates that distributional fidelity is the metric most sensitive to LLM stochasticity, and the one dimension where Mistral-7B's advantage over classical models is least consistent.

\subsection{Temporal Dynamics and Forecasting Accuracy}
\label{subsec:temporal}
\begin{figure*}[t]
\centering
\includegraphics[width=0.95\linewidth]{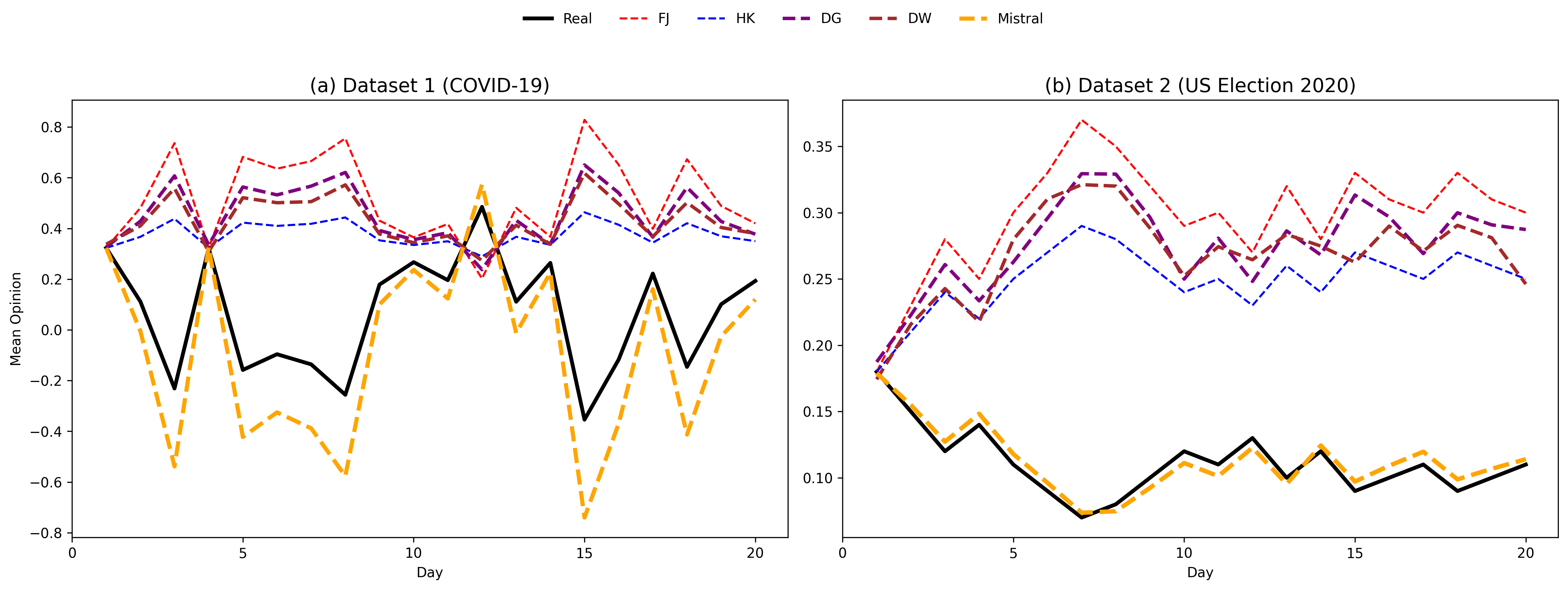}
\caption{Mean opinion trajectories over the validation period for the COVID-19 and US Election 2020 datasets.}
\label{fig:Meanopinion}
\end{figure*}

Fig.~\ref{fig:Meanopinion} shows mean opinion trajectories over the validation period. Mistral-7B closely tracks the empirical signal on both datasets, capturing gradual trends and short-term fluctuations alike.\\

On the COVID-19 dataset, the empirical trajectory exhibits pronounced irregular fluctuations consistent with the rapidly evolving public discourse driven by policy changes, news cycles, and emotionally charged communication. Mistral-7B tracks both the direction and magnitude of these changes, adapting agent responses to contextual exposure signals in a manner that reflects observed user behavior. Classical models produce substantially smoother trajectories: DeGroot and Friedkin-Johnsen converge toward consensus through repeated weighted averaging, whereas Hegselmann--Krause and Deffuant--Weisbuch stabilize into bounded clusters. All four classical models systematically underestimate empirical time variation, a structural limitation confirmed quantitatively by their high MAE and $\Delta\mathrm{Var}$ values in Table~\ref{tab:overall_fidelity}.

On the election dataset, the empirical trajectory displays more gradual directional shifts in mean opinion. Mistral-7B maintains alignment throughout the validation horizon, whereas classical models drift toward artificial equilibrium states that diverge from the observed sustained evolution of collective opinion.

\subsection{Ablation Study}
\label{subsec:ablation}
Table~\ref{tab:ablation_study} reports performance degradation when each framework component is removed. Three major results emerge consistently across both datasets which we discuss next.
 
\subsubsection{Agent attributes are the most critical component} 
Removing $\mathbf{a}_i$ produces the largest MAE degradation on both datasets: from 0.150 to 0.380 on COVID-19 (+0.230) and from 0.121 to 0.299 on the election dataset (+0.178). All other metrics degrade in parallel. Without persona, behavioral, and psychological heterogeneity, agents become effectively homogeneous and lose the ability to reproduce individual-level opinion variation. This confirms that empirically grounded attributes are the primary driver of Mistral-7B's simulation fidelity, not its linguistic generation capability alone.

\subsubsection{Memory and social exposure each contribute independently to digital twin fidelity} 
Removing memory $\mathcal{M}_i^t$ increases MAE to 0.265 and 0.216 on the COVID-19 and US Election 2020 datasets respectively. Removing social exposure $\mathcal{E}_i^t$ increases it to 0.255 and 0.195. Both degradations are consistent across all metrics and for both datasets. Memory enables longitudinal coherence: without it, agents respond to each time step in isolation, producing inconsistent trajectories. Social exposure enables collective dynamics: without it, agents evolve independently, and the network structure does not affect simulation outcomes.

\subsubsection{The component ranking is consistent across datasets}
For the MAE results, the ranking attributes $>$ memory $>$ exposure holds on both datasets. For secondary metrics, the ordering between memory and exposure is less stable: for the election dataset, removing memory causes only $+0.012$ degradation in $\Delta\mathrm{Var}$ whereas removing exposure causes $+0.073$, reversing the ranking for that metric. This indicates that memory's contribution is most pronounced at the individual prediction scale, while the social exposure's contribution is more uniformly distributed across evaluation metrics. All three components remain jointly necessary: removing any one of them significantly degrades performance across at least three of the four evaluation metrics.

\begin{table}[t]
\centering
\caption{Ablation study for proposed framework on COVID-19 and US Election 2020 datasets.}
\label{tab:ablation_study}

\renewcommand{\arraystretch}{1.1}
\setlength{\tabcolsep}{10pt}
\footnotesize

\begin{tabular}{lcccc}
\toprule
\multicolumn{5}{c}{\textbf{COVID-19 dataset}} \\

\textbf{Variant} & MAE & $\Delta r$ & $\Delta\mathrm{Var}$ & $\Delta\mathrm{EMD}$ \\
\midrule
Full model          & \textbf{0.150} & \textbf{0.120} & \textbf{0.106} & \textbf{0.293} \\
Remove attributes   & 0.380          & 0.201          & 0.290          & 0.485          \\
Remove memory       & 0.265          & 0.175          & 0.255          & 0.470          \\
Remove exposure    & 0.255          & 0.160          & 0.250          & 0.355          \\
\bottomrule
\end{tabular}

\vspace{0.8em}

\begin{tabular}{lcccc}
\toprule

\multicolumn{5}{c}{\textbf{US Election 2020 dataset}} \\

\textbf{Variant} & MAE & $\Delta r$ & $\Delta\mathrm{Var}$ & $\Delta\mathrm{EMD}$ \\
\midrule
Full Model          & \textbf{0.121} & \textbf{0.180} & \textbf{0.115} & \textbf{0.311} \\
Remove Attributes   & 0.299          & 0.410          & 0.258          & 0.495          \\
Remove Memory       & 0.216          & 0.223          & 0.127          &  0.392          \\
Remove Exposure    & 0.195          & 0.218          & 0.188          & 0.395          \\
\bottomrule
\end{tabular}

\end{table}

\subsection{Sensitivity Analysis}
\label{subsec:sensitivity}
\begin{figure*}[t]
\centering
\includegraphics[width=0.95\linewidth]{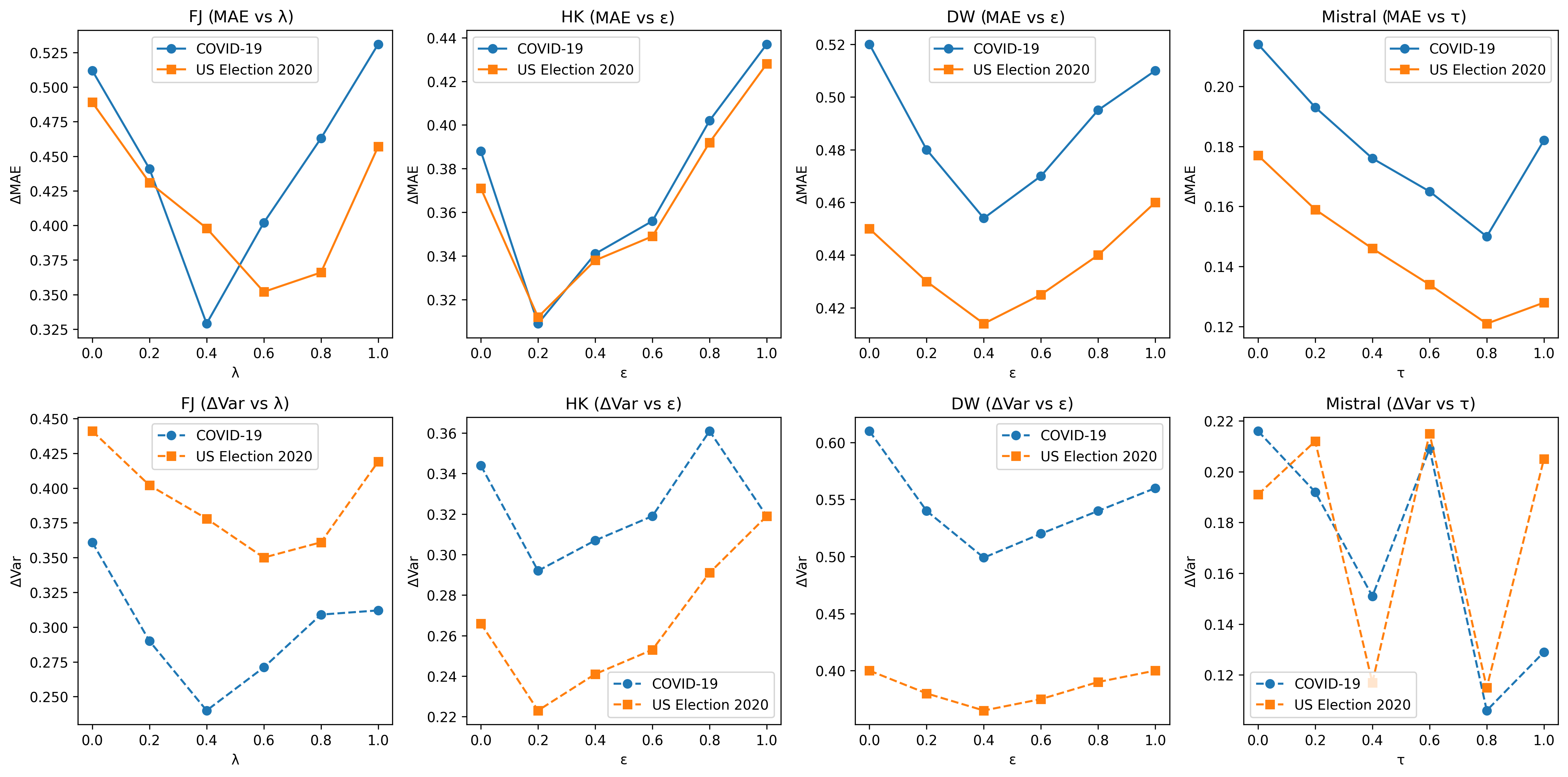}
\caption{Sensitivity of MAE (top row) and $\Delta \mathrm{Var}$ (bottom row) to hyperparameters for Friedkin--Johnsen (FJ), Hegselmann--Krause (HK), Deffuant--Weisbuch (DW), and Mistral across the COVID-19 and US Election 2020 datasets.}
\label{fig:sensitivity}
\end{figure*}

Table~\ref{tab:sensitivity_all_both} summarizes the sensitivity results of all four models to hyperparameter variation. Each row corresponds to a single configuration value, and the four evaluation metrics (MAE, $\Delta r$, $\Delta Var$, $\Delta EMD$) are computed independently for that configuration. Fig.~\ref{fig:sensitivity} visualizes the corresponding effects on MAE (top row) and $\Delta Var$ (bottom row).  The results reveal that classical models exhibit substantial hyperparameter sensitivity.

For the Friedkin--Johnsen model, the performance varies sharply with $\lambda$: moderate values (e.g., $\lambda=0.4$ for COVID-19 and $\lambda=0.6$ for the U.S. Election dataset) yield the lowest discrepancies in Table~\ref{tab:sensitivity_all_both}. This pattern is directly reflected in Fig.~\ref{fig:sensitivity}, where the MAE curve exhibits a clear minimum at these mid-range settings. Extreme values either collapse the model into DeGroot-like averaging ($\lambda=0$) or suppress opinion updating ($\lambda=1$), producing the large MAE and $\Delta Var$ spikes visible at the edges of the Friedkin--Johnsen curves in Fig.~\ref{fig:sensitivity}.

Hegselmann--Krause and Deffuant--Weisbuch models display similar instability with respect to the confidence bound $\epsilon$: Table~\ref{tab:sensitivity_all_both} shows that small $\epsilon$ values fragment the network and inflate all discrepancy metrics, whereas large $\epsilon$ values induce premature convergence. Fig.~\ref{fig:sensitivity} mirrors this behavior through steep gradients in both MAE and $\Delta Var$ as $\epsilon$ moves away from the narrow region where performance is optimal. No single configuration generalizes across datasets for any classical model, and the irregular, non-smooth shapes of their curves in Fig.~\ref{fig:sensitivity} reinforce the lack of robustness indicated numerically in Table~\ref{tab:sensitivity_all_both}.

In contrast, Mistral-7B exhibits consistent stability across all temperature settings. Table~\ref{tab:sensitivity_all_both} shows only minor variation in MAE, $\Delta Var$, and $\Delta EMD$ across $\tau$, and Fig.~\ref{fig:sensitivity} confirms this through the smooth, shallow curves that remain nearly flat across the entire temperature range. The best performance consistently occurs at $\tau=0.8$, where both MAE and $\Delta Var$ reach their minima, matching the lowest-discrepancy entries in Table~V. Lower temperatures slightly increase MAE by reducing generative diversity, while higher temperatures introduce additional stochasticity without improving fidelity. However, these effects remain small relative to the dramatic fluctuations observed in the classical models. The alignment between the narrow performance band in Table~V and the stable curves in Fig.~\ref{fig:sensitivity} demonstrates that the digital twin’s predictive advantage arises from attribute-conditioned generative reasoning rather than from precise hyperparameter tuning, and that this robustness holds for both datasets.

\begin{table*}[t]
\centering
\caption{Sensitivity analysis under hyperparameter variation.}
\label{tab:sensitivity_all_both}

\setlength{\tabcolsep}{6pt}
\renewcommand{\arraystretch}{1.12}
\footnotesize
\newcolumntype{C}{>{\centering\arraybackslash}p{0.06\textwidth}}

\resizebox{\textwidth}{!}{%
\begin{tabular}{|C C C C C
!{\vrule width 1.3pt}
C C C C C
!{\vrule width 1.3pt}
C C C C C
!{\vrule width 1.3pt}
C C C C C|}
\specialrule{1.3pt}{0pt}{0pt}

\multicolumn{5}{c!{\vrule width 1.3pt}}{\textbf{Friedkin--Johnsen ($\lambda$)}} &
\multicolumn{5}{c!{\vrule width 1.3pt}}{\textbf{Hegselmann--Krause ($\epsilon$)}} &
\multicolumn{5}{c!{\vrule width 1.3pt}}{\textbf{Mistral ($\tau$)}} &
\multicolumn{5}{c|}{\textbf{Deffuant--Weisbuch  ($\epsilon$)}} \\
\specialrule{0.9pt}{2pt}{2pt}

$\lambda$ & MAE  & $\Delta r$ & $\Delta\mathrm{Var}$ & $\Delta \mathrm{EMD}$ &
$\epsilon$ & MAE & $\Delta r$ & $\Delta\mathrm{Var}$ & $\Delta \mathrm{EMD}$ &
$\tau$ & MAE & $\Delta r$ & $\Delta\mathrm{Var}$ & $\Delta \mathrm{EMD}$ &
$\epsilon$ & MAE & $\Delta r$ & $\Delta\mathrm{Var}$ & $\Delta \mathrm{EMD}$ \\
\specialrule{1.0pt}{2pt}{2pt}

\multicolumn{20}{c}{\textbf{ COVID-19 dataset}} \\
\specialrule{0.8pt}{2pt}{2pt}

0.0 & 0.512 & 0.389 & 0.361 & 0.344 &
0.0 & 0.388 & 0.366 & 0.344 & 0.381 &
0.0 & 0.214 & 0.188 & 0.216 & 0.371 &
0.0 & 0.520 & 0.450 & 0.610 & 0.480 \\

0.2 & 0.441 & 0.333 & 0.290 & 0.392 &
\textbf{0.2} & \textbf{0.309} & \textbf{0.303} & \textbf{0.292} & \textbf{0.328} &
0.2 & 0.193 & 0.184 & 0.192 & 0.358 &
0.2 & 0.480 & 0.380 & 0.540 & 0.450 \\

\textbf{0.4} & \textbf{0.329} & \textbf{0.250} & \textbf{0.240} & \textbf{0.311} &
0.4 & 0.341 & 0.319 & 0.307 & 0.359 &
0.4 & 0.176 & 0.212 & 0.151 & 0.349 &
\textbf{0.4} & \textbf{0.454} & \textbf{0.342} & \textbf{0.499} & \textbf{0.411} \\

0.6 & 0.402 & 0.301 & 0.271 & 0.335 &
0.6 & 0.356 & 0.333 & 0.319 & 0.374 &
0.6 & 0.165 & 0.201 & 0.209 & 0.341 &
0.6 & 0.470 & 0.360 & 0.520 & 0.430 \\

0.8 & 0.463 & 0.352 & 0.309 & 0.317 &
0.8 & 0.402 & 0.387 & 0.361 & 0.421 &
\textbf{0.8} & \textbf{0.150} & \textbf{0.120} & \textbf{0.106} & \textbf{0.293} &
0.8 & 0.495 & 0.390 & 0.540 & 0.455 \\

1.0 & 0.531 & 0.421 & 0.312 & 0.402 &
1.0 & 0.437 & 0.312 & 0.319 & 0.395 &
1.0 & 0.182 & 0.189 & 0.129 & 0.455 &
1.0 & 0.510 & 0.410 & 0.560 & 0.470 \\

\specialrule{1.2pt}{2pt}{2pt}
\multicolumn{20}{c}{\textbf{US Election 2020 dataset}} \\
\specialrule{0.8pt}{2pt}{2pt}

0.0 & 0.489 & 0.502 & 0.441 & 0.471 &
0.0 & 0.371 & 0.387 & 0.266 & 0.402 &
0.0 & 0.177 & 0.216 & 0.191 & 0.461 &
0.0 & 0.450 & 0.520 & 0.400 & 0.390 \\

0.2 & 0.431 & 0.564 & 0.402 & 0.498 &
\textbf{0.2} & \textbf{0.312} & \textbf{0.323} & \textbf{0.223} & \textbf{0.337} &
0.2 & 0.159 & 0.213 & 0.212 & 0.352 &
0.2 & 0.430 & 0.500 & 0.380 & 0.370 \\

0.4 & 0.398 & 0.531 & 0.378 & 0.476 &
0.4 & 0.338 & 0.342 & 0.241 & 0.361 &
0.4 & 0.146 & 0.251 & 0.117 & 0.344 &
\textbf{0.4} & \textbf{0.414} & \textbf{0.493} & \textbf{0.365} & \textbf{0.358} \\

\textbf{0.6} & \textbf{0.352} & \textbf{0.497} & \textbf{0.350} & \textbf{0.442} &
0.6 & 0.349 & 0.356 & 0.253 & 0.374 &
0.6 & 0.134 & 0.189 & 0.215 & 0.339 &
0.6 & 0.425 & 0.505 & 0.375 & 0.365 \\

0.8 & 0.366 & 0.509 & 0.361 & 0.455 &
0.8 & 0.392 & 0.401 & 0.291 & 0.419 &
\textbf{0.8} & \textbf{0.121} & \textbf{0.180} & \textbf{0.115} & \textbf{0.311} &
0.8 & 0.440 & 0.515 & 0.390 & 0.380 \\

1.0 & 0.457 & 0.581 & 0.419 & 0.519 &
1.0 & 0.428 & 0.437 & 0.319 & 0.456 &
1.0 & 0.128 & 0.208 & 0.205 & 0.394 &
1.0 & 0.460 & 0.530 & 0.400 & 0.395 \\

\specialrule{1.3pt}{0pt}{0pt}
\end{tabular}%
}
\end{table*} 

\section{Discussion}
\label{sec:Discussion}

\subsection{Why Mistral Reproduces Empirical Opinion Dynamics}
\label{subsec:why}
Three mechanisms jointly explain Mistral's predictive advantage over classical models. First, belief updating in real social networks involves linguistic interpretation and contextual inference processes that fixed numerical aggregation rules structurally cannot capture. Second, empirically derived agent attributes introduce the behavioral and psychological heterogeneity that drives individual-level variation in opinion trajectories. This aligns with Park et al.~\cite{park2024generative}, who showed that persona- grounded agents replicate individual attitudes with substantially higher fidelity than generic simulations. Third, persistent memory allows agents to maintain longitudinal coherence across the validation horizon, producing trajectories that reflect prior belief states rather than reacting independently to each timestep. Classical models lack all three mechanisms, which explains why they systematically converge toward smooth equilibrium whereas empirical trajectories exhibit abrupt, irregular fluctuations.

\subsection{The Role of Empirical Grounding}
\label{subsec:grounding}
The ablation study confirms that empirical grounding, not linguistic generation capability alone, drives simulation fidelity. Removing agent attributes causes the largest degradation across all metrics and both datasets, demonstrating that Mistral's advantage depends critically on being conditioned on real user characteristics rather than operating as a generic language model. Our results show that the Mistral model does not collapse to consensus, $\Delta\mathrm{Var}$ remains low for both datasets, suggesting that attribute conditioning and stubbornness estimation introduce sufficient heterogeneity to counteract this tendency, though the precise mechanism warrants further investigation.

\subsection{Mistral as a Generative Digital Twin Engine} \label{subsec:engine}
Beyond practical utility, these results reveal a theoretical implication. Mistral-7B was not trained for opinion dynamics prediction. Its fidelity arises because large-scale pretraining encodes implicit models of social reasoning and belief revision that become predictive when anchored to empirical network structure and user attributes that function as a cognitive surrogate activated by real interaction data rather than synthetic environments. This framing aligns with Larooij and T\"{o}rnberg~\cite{larooij2025validation}, who called for simulations grounded in empirical data rather than subjective believability, and suggest that the value of LLMs in social simulation lies not in linguistic fluency alone but in their capacity to serve as structured knowledge bases for human behavioral reasoning when empirically conditioned.

\subsection{Implications and Generalization} \label{subsec:implications}
Mistral's ability to reproduce empirical opinion trajectories in a cloned interaction network opens concrete pathways for scenario-based social simulation. Researchers can use this framework to evaluate the potential impact of communication interventions, information campaigns, or structural network changes before real-world deployment. Our proposed framework provides capabilities that classical models cannot provide, due to their fixed update rules.

\section{Limitations and Future Research Directions}
\label{sec:limitations}
Despite its strong performance, our proposed framework has several limitations that we acknowledged. 

\begin{itemize}

\item {\em Datasets:} We conducted the empirical evaluation on two Twitter-based datasets covering COVID-19 discourse and the 2020 U.S.\ presidential election. While these provide distinct socio-political contexts, the limited scope restricts generalization to other platforms, languages, or topic domains. Differences in interaction norms, content modality, and network topology may affect simulation fidelity in ways not captured here. To address this limitation, as part of our future work, we will evaluate our proposed approach across diverse settings to establish the boundary conditions of the framework's applicability.

\item {\em LLM:} We evaluated the proposed framework using a single open-source LLM (Mistral-7B). Simulation fidelity may vary across models with different pretraining corpora, alignment strategies, or parameter scales. Results obtained with Mistral-7B should not be assumed to generalize to other LLMs without empirical verification, and comparative multi-model evaluation represents a direct next step. To address this issue, in the future, we will evaluate the framework across multiple open-source LLMs of different scales and alignment strategies, to determine which model characteristics most affect simulation fidelity and to establish the robustness of our findings beyond Mistral-7B.

\item{\em Static network structure:} The construction of agent attributes relies on features such as sentiment scores, rhetorical characteristics, and network statistics extracted from social media data, while assuming a fixed network structure during simulation. Although these attributes provide useful approximations of user behavior, real-world social networks evolve as users form new connections and disengage over time. To address this limitation, as part of our future work, we will  consider dynamic network evolution that will allow the framework to capture the co-evolution of interaction topology and opinion dynamics. One example could be the modeling how opinion-driven homophily reshapes the network during a health crisis or election cycle.

\item {\em Opinion Representation:} The opinion trajectories used for calibration and evaluation are derived from sentiment polarity rather than from direct stance annotation. While this proxy is reasonable for topics such as vaccination and electoral preference, where affective tone and stance are strongly correlated, it does not capture cases of sarcasm, stance-incongruent affect (e.g., anger expressed in defense of a position), or topics where sentiment and stance are more loosely coupled. Future work should validate this proxy against human-annotated stance labels.

\item{\em Computation cost:} Finally, our approach introduces a higher computational cost compared with classical models due to the reliance on LLM inference at each simulation step. While the experiments involve networks of several hundred agents and remain computationally tractable, real-world social platforms often consist of thousands or millions of users. Scaling the framework to such settings will require additional optimization strategies that we will also investigate in the future to address the scalability issue.

\end{itemize}

\section{Conclusion}
\label{sec:Conclusion}
In this work, we proposed a generative digital twin framework that clones empirical social interaction networks to predict opinion evolution through the Mistral model that is conditioned on multi-dimensional user attributes, persistent memory, and social exposure. We evaluated the proposed framework on two real-world Twitter datasets. The framework demonstrated the ability to reproduce opinion trajectories that fixed numerical update rules systematically fail to capture. Specifically, it reduces individual-level prediction error by more than 50\% over the strongest classical baseline and achieves consistent improvements across polarization, structural, and distributional evaluation scales. Additionally, the ablation experiments confirm that empirically grounded agent attributes are the primary driver of this fidelity, demonstrating that the Mistral model functions as a predictive social cognition surrogate when anchored to real network structure rather than synthetic environments. Extending validation to additional platforms, languages, and topic domains beyond Twitter represents the most direct path toward establishing the generalizability of this simulation paradigm and its applicability to a broader range of social contexts. This will be the future research direction we will undertake next.

\appendices
\section{List of Acronyms}

\begin{table}[h]
\centering
\renewcommand{\arraystretch}{1.3}
\begin{tabular}{ll}
\hline
\textbf{Acronym} & \textbf{Description} \\
\hline
\textbf{AI}   & Artificial intelligence \\
\textbf{DG}   & DeGroot \\
\textbf{DW}   & Deffuant--Weisbuch \\
\textbf{EMD}  & Earth Mover's distance \\
\textbf{FJ}   & Friedkin--Johnsen \\
\textbf{HK}   & Hegselmann--Krause \\
\textbf{LLM}  & Large language model \\
\textbf{MAE}  & Mean absolute error \\
\textbf{ABM}  & Agent-Based Modeling  \\

\hline
\end{tabular}
\end{table}




%



\bibliographystyle{IEEEtran}
\bibliography{references}
\end{document}